\documentclass[copyright,creativecommons]{eptcs}
\providecommand{\event}{ICLP 2024} 

\usepackage{iftex}

\ifpdf
  \usepackage{underscore}         
  \usepackage[T1]{fontenc}        
\else
  \usepackage{breakurl}           
\fi

\usepackage{graphicx} 
\usepackage{amsmath}
\usepackage{multirow}
\usepackage{hyperref}
\usepackage{subcaption}
\usepackage{multirow}

\usepackage[T1]{fontenc}
\usepackage{alltt}
\usepackage{amsmath,amssymb}
\usepackage{underscore}

\newcommand{\Vex}[1]{\vspace{#1ex}}
\newenvironment{code}{\Vex{-.5}\begin{alltt}\footnotesize}{\end{alltt}\Vex{-.5}}
\NewDocumentCommand{\co}{+m}{\mbox{\small\tt #1}} 
\newenvironment{example}{\Vex{.5}\par\textbf{\textit{Example}}.}{\hfill$\blacksquare$\par}
\newcommand{\mypar}[1]{\Vex{-2}\paragraph{\bf #1.~}}
\newcommand{\mysec}[1]{\Vex{-1}\section{#1}\Vex{-.5}}
\newcommand{\mysubsec}[1]{\Vex{-1}\subsection{#1}\Vex{-.5}}

\title{LP-LM: No Hallucinations in Question Answering with Logic Programming\Vex{-1}}

\author{Katherine Wu\footnote{Work done as a student at Stony Brook University}
\institute{Cornell University} 
\email{kaw324@cornell.edu}
\and
Yanhong A. Liu
\institute{Stony Brook University} 
\email{liu@cs.stonybrook.edu}}

\def\titlerunning{LP-LM}
\def\authorrunning{Katherine Wu \& Yanhong A. Liu}
\begin{document}
\maketitle

\Vex{-2}
\begin{abstract}
Large language models (LLMs) are able to generate human-like responses to user queries. However, LLMs exhibit inherent limitations, especially because they hallucinate. This paper introduces LP-LM, a system that grounds answers to questions in known facts contained in a knowledge base (KB), facilitated through semantic parsing in Prolog, and always produces answers that are reliable. 

LP-LM generates a most probable constituency parse tree along with a corresponding Prolog term for an input question via Prolog definite clause grammar (DCG) parsing. The term is then executed against a KB of natural language sentences also represented as Prolog terms for question answering. 
By leveraging DCG and tabling, LP-LM runs in 
    linear time in the size of input sentences for sufficiently many grammar rules.
Performing experiments comparing LP-LM with current well-known LLMs in accuracy, we show that LLMs hallucinate on even simple questions, unlike LP-LM.
\end{abstract}

\mysec{Introduction}
Large language models (LLMs) hallucinate, i.e., generate information that appears plausible but is factually incorrect~\cite{hallucination-survey}. 
This unfortunately poses a challenge to question answering tasks, as users desire reliable answers given a query, but hallucination misleads users and erodes the system reputation~\cite{chatgpt-qas}.
To overcome this challenge, better retrieval models that retrieve relevant information according to queries as well as better generation models that synthesize more accurate answers from knowledge sources are needed. This paper sheds light on how logic programming can be used to push progress on the former. We describe LP-LM, a system that considers the structure of natural language sentences when retrieving answers to user queries. Unlike LLMs, which are pre-trained so that for any given input the statistically best matching output based on its training is given, LP-LM seeks to answer questions in a logical and verifiable way via matching and substitution of facts. 

We use probabilistic context-free grammar (PCFG) productions to model the structures of valid English sentences and create a knowledge base (KB) consisting of English sentences represented as Prolog terms. The term structure models relationships between entities in sentences precisely. When the user asks a natural language question, LP-LM generates the most probable constituency parse tree of the input sentence, translates the parse tree into a corresponding Prolog term for knowledge representation, and then matches the term against the KB of Prolog terms to retrieve an answer using unification. Utilizing Prolog's definite clause grammar (DCG) and tabling in our implementation, LP-LM proves to be extremely efficient, especially for grammars with a significant number of production rules. We have implemented LP-LM using the Prolog system XSB~\cite{SagSW94xsb,xsb22}, and our implementation is publicly available.\footnote{\url{https://github.com/katherinewu312/lp-lm}} 

The rest of the paper is organized as follows. Section~\ref{sec:background} defines terms used throughout the paper. Section~\ref{sec:experiments} compares LP-LM with current LLMs by highlighting simple example problems on which current LLMs fail but LP-LM succeeds. 
Section~\ref{sec:unification} describes how LP-LM works, giving an example of an execution along with the underlying details of the execution.
Section~\ref{sec:related} discusses related work and concludes.

\mysec{Background}
\label{sec:background}

We introduce probabilistic context-free grammars and key logic programming features used.

\mypar{Probabilistic context-free grammar}

A probabilistic context-free grammar (PCFG) is a formal grammar used in natural language processing and computational linguistics~\cite{pcfg-1,pcfg-2}. PCFGs 
associate
probabilities 
with
the production rules of the grammar. These probabilities reflect the likelihood of a particular rule being used in generating or deriving a sentence. For any non-terminal in a PCFG, the probabilities associated with rules corresponding to that non-terminal must sum to 1. 

PCFGs are essential for capturing the ambiguity of natural language, and are particularly useful in tasks such as syntactic parsing, which uses dynamic programming algorithms to compute the most likely parse tree of a sentence given a statistical model of the syntactic structure of the language. The Cocke-Younger-Kasami algorithm (CYK) (Cocke 1969~\cite{cocke-cyk}; Younger 1967~\cite{younger-cyk}; Kasami 1965~\cite{kasami-cyk}), the Earley algorithm~\cite{earley}, and the shift-reduce algorithm~\cite{shift-reduce} are at the core of most common algorithms for natural language parsing, both constituency-based and dependency-based. 


\mypar{Definite clause grammar} Definite Clause Grammars (DCGs) are a convenient way to represent grammatical relationships for parsing applications. They can be used to progressively build a parse tree as grammar rules are applied. DCG provides a syntax for writing more readable grammar parsing rules, and the DCG preprocessor is able to translate a DCG rule into pure Prolog. The arrow operator indicates a DCG rule, which replaces the normal neck ``\co{:-}'' used in Prolog clauses, and square brackets are used to indicate terminal symbols of the grammar. Figure~\ref{fig:dcg-example-1} gives an example. Works similar to DCGs include stochastic DCGs~\cite{stochastic-dcg}, relaxed unification grammars~\cite{uni-grammars-relaxed}, and probabilistic unification grammars~\cite{uni-grammars-prob}.

\begin{figure*}[t!]\small
    \centering
    \begin{subfigure}[t]{0.5\textwidth}
        \centering
        \begin{code}
        s --> np, vp.
        np --> dt, nn.
        np --> nn.
        vp --> vi.
        dt --> [the].
        nn --> [man].
        vi --> [sleeps].
    
        ?- s([the,man,sleeps],[]).
        yes
        \end{code}
    \end{subfigure}%
    ~ 
    \begin{subfigure}[t]{0.5\textwidth}
        \centering
        \begin{code}
        s(A,B) :- np(A,C), vp(C,B).
        np(A,B) :- dt(A,C), nn(C,B).
        np(A,B) :- nn(A,B).
        vp(A,B) :- vi(A,B).
        dt([the|R],R).
        nn([man|R],R).
        vi([sleeps|R],R).
    
        ?- s([the,man,sleeps],[]).
        yes
        \end{code}
    \end{subfigure}\Vex{-3}
    \caption{An example Prolog DCG and a parse. The two Prolog versions are equivalent.}
    \label{fig:dcg-example-1}
\end{figure*}

\mypar{Tabling} Tabling consists of maintaining a table of goals that are called during execution, along with their answers, and then using the answers directly when the same goal is subsequently called. The idea is to never evaluate the same call twice. It helps improve the running time drastically, including terminating efficiently in situations where Prolog goes into an infinite loop following the same calls repeatedly.
%

\mypar{Unification} The way in which Prolog matches two terms is called unification. For example, applying unification of \co{foo(a,X)} and \co{foo(Y,b)}: the principal functor of both terms is \co{foo}; the arguments of \co{foo(a,X)} are \co{a} and \co{X}, the arguments of \co{foo(Y,b)} are \co{Y} and \co{b}; so \co{a} and \co{Y} must unify, instantiating \co{Y} to \co{a}, and \co{X} and \co{b} must unify, instantiating \co{X} to \co{b}; and finally the resulting term after unification is \co{foo(a,b)}.

\mysec{Comparison with existing LLMs}
\label{sec:experiments}
Before delving into the key designs of LP-LM, we first compare our system with existing LLMs to highlight the motivation behind our work. We focus on the following well-known models: GPT-4o, GPT-4o mini, and Gemini. In particular, we show that the context-awareness of these LLMs are actually quite poor in question answering tasks, and that the LLMs struggle to perform tasks involving even single facts, thus limiting their potential to complete more complex reasoning tasks.  


Table~\ref{table:hallucination-prompts} illustrates the comparisons. The answers shown are from the first run of the models. Note that for the first two examples given, the inputs are entered independently, and we only show the answer that corresponds to the last input due to space. The last two examples consider the separate inputs from the earlier examples as one prompt, but even with this the models still hallucinate. 
The examples demonstrate that current LLMs exhibit a lack of understanding and ability to reason about the relationships between different concepts and entities, and are only able to generate text based on statistical correlations they have learned from their training data.

    

\begin{table}[tp]
\centering
\begin{tabular}{| @{~} p{2.7cm}||@{~}p{1.9cm}|p{3.2cm}|p{3.1cm}|p{3cm}|  }
\hline

 Prompt Input \newline from User & LP-LM\newline Answer & ChatGPT Answer\newline using GPT-4o & ChatGPT Answer\newline using GPT-4o mini& Gemini Answer\\
 

 \hline \hline
 ``Furosemide causes temporary hearing loss.'' 
 \newline
 
 ``What causes temporary hearing loss?'' &
 ``Furosemide causes temporary hearing loss.'' & 
 ``Temporary hearing loss can be caused by a variety of factors. Some of the common causes include: earwax blockage, ear infections, noise exposure, ...''
 &
 ``Temporary hearing loss can result from various factors, including: exposure to loud noise, ear infections, earwax blockage, ...'' &  
 
 ``Several factors can cause temporary hearing loss: ototoxicity, noise exposure, earwax buildup, Meniere's disease, ...'' \\
 \hline
 
 ``Fir trees can grow in human lungs.''
 \newline
 
 ``What can grow in human lungs?''
 &
 ``Fir trees can grow in human lungs.''  & 
 
 ``Several abnormal growths or conditions can occur in human lungs, typically as a result of disease, infection, or exposure to harmful substances...'' 
 &
 ``In human lungs, the following can grow or develop: tumors, infections, cysts, ...'' &
 
 ``Only a few things can grow in human lungs, and they are all extremely rare and unusual... fungal infections, bacteria, parasites...'' \\
 
 \hline
 \hline
 ``Furosemide causes temporary hearing loss. What causes temporary hearing loss?'' &
 ``Furosemide causes temporary hearing loss.'' & 
 ``Temporary hearing loss can be caused by a variety of factors, including: exposure to loud noises, ear infection medications, sudden changes in air pressure...''
 &
 ``Temporary hearing loss can be caused by various factors, including: medications, exposure to loud noise, ear infections...'' &  
 
 ``Furosemide can cause temporary hearing loss due to a condition called ototoxicity. This occurs when a medication damages the inner ear, which is responsible for hearing.'' \\

 \hline
 ``Fir trees can grow in human lungs. What can grow in human lungs?''&

 ``Fir trees can grow in human lungs.'' &
 
 ``Infections, tumors, or foreign objects can be found in the lungs, but trees, plants, or other organisms generally cannot grow there...'' 
 &
 ``In reality, fir trees cannot grow in human lungs. However, certain things can grow or develop in human lungs, including tumors, infections, cysts, ...'' &  
 
 ``Fir trees cannot grow in human lungs...'' \\
 
 \hline
\end{tabular}
\caption{Table comparing LP-LM with existing LLMs for example prompts}
\label{table:hallucination-prompts}
\end{table}

\mysec{Executing LP-LM using Prolog unification}
\label{sec:unification}
We outline a typical LP-LM workflow here and then give an example. LP-LM's KB of Prolog terms is used to provide context. 

\begin{itemize}
    \item \textbf{Input}: A user's input can be either a \textit{statement} (e.g., ``suppose I say the black bird flies bravely.'') which eventually leads to a question, or a \textit{question} (e.g., ``how does the black bird fly?'') following some previous statement. If there are multiple sentences in the input, they are processed one at a time.

    \item \textbf{Retrieval from or insertion into KB}: The input is parsed using Prolog DCG rules, and a constituency tree and associated Prolog term is generated from the parser. For statements, insertions into the KB are performed: the term is inserted dynamically into the KB. For questions, retrievals from the KB are performed: the term is matched against the KB and an answer is obtained by unification. 

    \item \textbf{Post-processing}: Optionally, the results can be translated to a natural language answer. 
\end{itemize}

We show an example of an LP-LM execution, after which we describe the internal steps of the retrieval and insertion process.


\begin{example}
Consider an example sentence that includes a determiner, adjective, noun, verb, and adverb. This statement gets inserted into the specialized KB of Prolog terms via the predicate \co{add\_kb}:

\begin{code}
    ?- add_kb(`the black bird flies bravely').
\end{code}
After statements, one can perform queries, which can either be yes/no or wh- questions, where predicate \co{query\_kb} does the query.

\begin{code}
    ?- query_kb(`how does the black bird fly').
    Answer: bravely

    ?- query_kb(`who flies bravely').
    Answer: black(bird)

    ?- query_kb(`does the black bird fly bravely').
    Answer: yes
\end{code}
One can also remove previous statements as follows, where predicate \co{remove\_kb} does the removal:    
\begin{code}
    ?- remove_kb(`the black bird flies bravely').
\end{code}\Vex{-2.5}
\end{example}

LP-LM takes into account the various verb tenses in the English language: simple, perfect, continuous, and perfect continuous tenses, each with their own past, present, and future tenses. Additionally, LP-LM supports many sentence patterns. These current patterns encompass the prominent structures of simple declarative sentences in English, and adding more patterns to the system for generalization purposes is straightforward. Regardless of the sentence, an English sentence will always have two parts: a subject and a verb. When generating the Prolog term for a given sentence, the root form of the verb is always used as the functor. More details are described in our implementation.


\mysubsec{Insertions into KB}
With non-queries, or what we call statements, insertions into the KB are done. A tokenizer is first used to extract out each word in the statement, then a top-down evaluation method is used to generate the parse tree and Prolog term for the sentence. The Prolog term is added to the KB. We take the basic sentence, ``Bob runs''. The DCG rules are applied in the following order:
\begin{enumerate}

\item The DCG rule
\begin{code}
    s(s(NP,VP),Sem,P) --> np(NP,X,P1), vp(VP,Y,_,P2), \{Sem=..[Y,X]\}, \{P is P1*P2*0.25\}.
\end{code}
is first matched with the sentence. Variable \co{Sem} represents the Prolog term, where \co{Y} is the functor of the term and \co{X} is the argument, which is generated incrementally as the words in the input sentence are matched to a DCG rule one by one.

\item The DCG rule 
\begin{code}
    np(np(PN),X,P) --> pn(PN,X,P1), \{P is P1*0.2\}.
\end{code}
is matched next, followed by the DCG rule 
\begin{code}
    pn(pn(X),X,1.0)	--> [X], \{pronoun(X)\}.
\end{code}
which checks if ``Bob'' is a pronoun, as the variable \co{X} represents ``Bob''.

\item The DCG rule 
\begin{code}
    vp(vp(VB),Verb,C,P) --> v(VB,Verb,C,P1), \{P is P1*0.09\}.
\end{code}
is matched next, followed by the DCG rule
\begin{code}
    v(v(X),Vx,C,1.0) --> [X], \{verb(Vx,C,[X],[])\}.
\end{code}
which checks if ``runs'' is a verb, as the variable \co{X} represents ``runs''.
\item The Prolog term \co{runs(Bob)} is obtained, with the parse tree \co{s(np(pn(Bob)),vp(v(runs)))}, with probability 0.0045. This is the most probable parse tree. The term is added to the KB.

\end{enumerate}

\mysubsec{Retrievals from KB}
With queries, retrievals from the KB are done. The parse tree and Prolog term for the question is generated the same way. The resulting term is then matched against the KB of terms, and unification is used to obtain the answer to the question. Consider the question ``who runs'', which should return the answer ``Bob'' per the example above. The DCG rules are applied as follows:
\begin{enumerate}

\item The DCG rule
\begin{code}
    q(q(QW,VB), X, P) --> qw(QW,_Qw,P1), v(VB,Verb,_,P2), 
        \{Sem=..[Verb,X],Sem\}, \{P is P1*P2*0.05\}.
\end{code}
is applied, where \co{qw} represents the question word ``who'' and \co{v} represents the verb ``runs''.

\item The DCG rule 
\begin{code}
    qw(qw(X),X,1.0) -->[X], \{qword(X)\}.
\end{code}
is matched next, which checks if ``who'' is a question word, as the variable \co{X} represents ``who''.

\item The DCG rule 
\begin{code}
    v(v(X),Vx,C,1.0) -->[X], \{verb(Vx,C,[X],[])\}.
\end{code}
is matched next, which checks if ``runs'' is a verb, as the variable \co{X} represents ``runs''.

\item The Prolog term \co{run(X)} is obtained, along with the associated parse tree of \co{q(qw(who),v(runs))} with probability 0.05, the most probable tree. The term \co{run(X)}, where \co{X} is a variable, will be unified with a matching rule in the KB, which in this case is \co{run(Bob)}. Thus, \co{X} = \co{Bob}. 

\end{enumerate}
For yes/no questions such as ``does Bob run?'', the tree is \co{q(av(does),np(pn(bob)),v(run))} and the Prolog term generated is thus \co{run(bob)}. In this case, LP-LM checks if there is an exact match of this term in the KB and a true/false answer is returned by the Prolog engine. 

\mysubsec{A note on DCG parsing efficiency}

To find the most probable parse tree in LP-LM, all possible parses of input segments that can contribute to the maximum probability are considered and compared, from which the parse with the maximum probability is constructed and returned. Despite this global optimality, the parsing that underlies LP-LM still proves to be efficient due to our use of Prolog DCGs and tabling. We have performed experiments testing the efficiency of DCGs and have shown that DCGs still outperform state-of-the-art bottom-up greedy parsing algorithms.

We evaluated DCG parsers on a total of 12 PCFGs: 3 left-recursive grammars, 3 right-recursive grammars, 3 unambiguous grammars, and 3 ambiguous grammars. For each type of grammar, we increase the size complexity by increasing the number of production rules with each test: the first test consisted of a trivial grammar with 3-10 production rules, the second test consisted of a more complex grammar with 20-50 production rules, and the third test consisted of the longest and most complex grammar with 100+ production rules. Within each test, 3-5 input sentences of increasing length satisfying the corresponding grammar were parsed, and the time of each parse recorded.

\begin{figure}[htp]
\centering
\includegraphics[width=.3\textwidth]{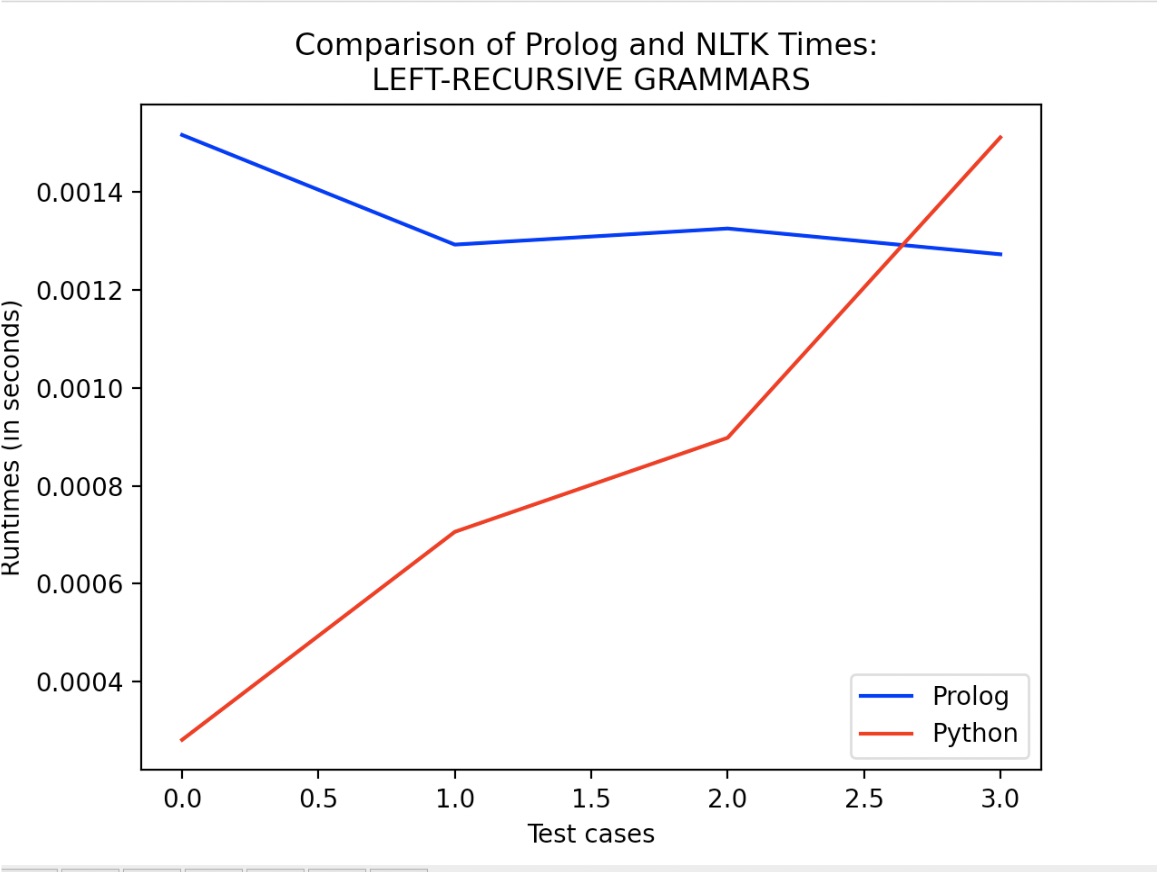}\hfill
\includegraphics[width=.3\textwidth]{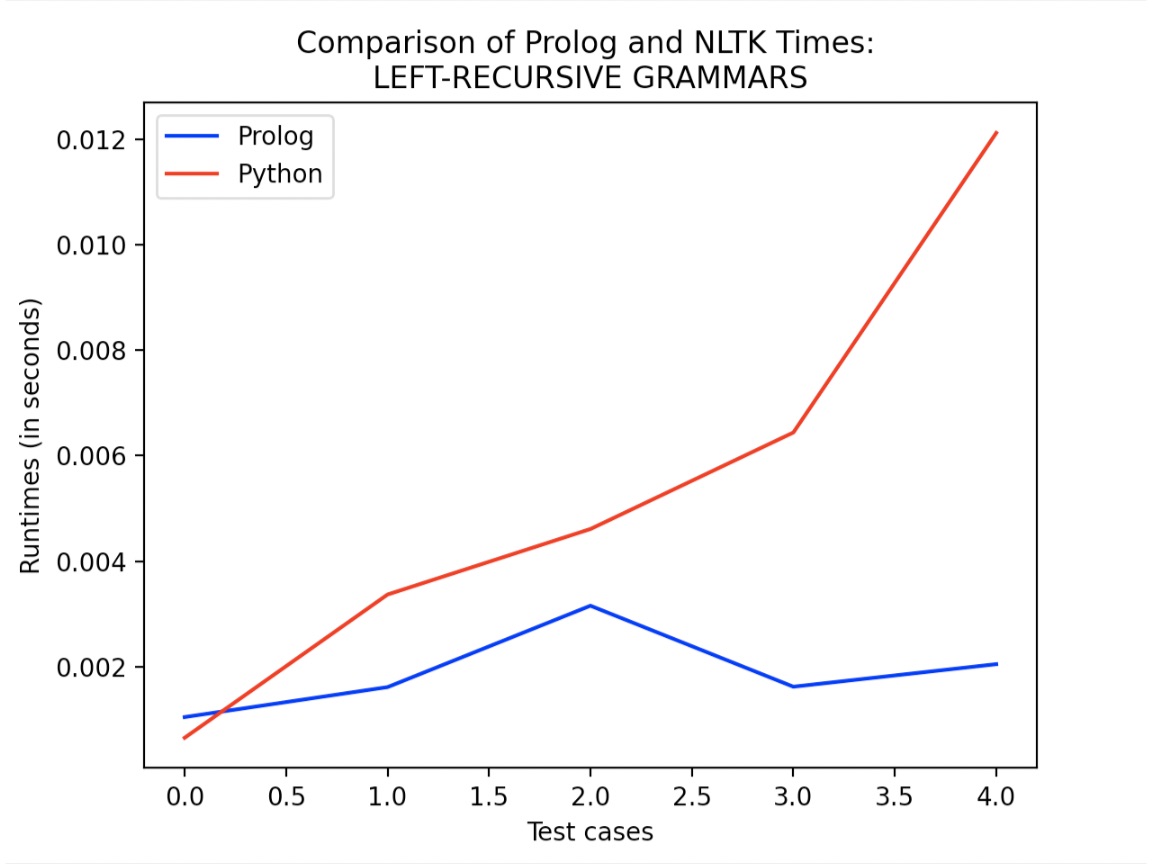}\hfill
\includegraphics[width=.3\textwidth]{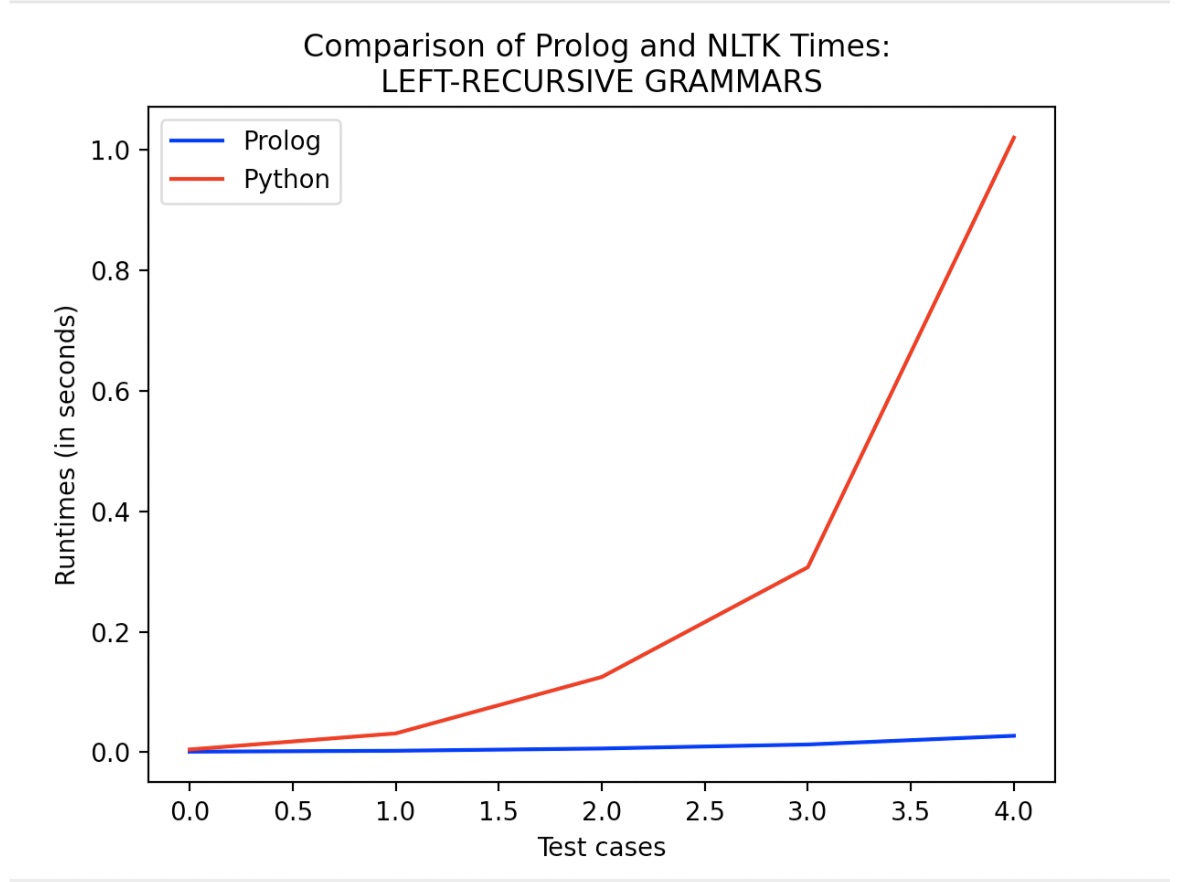}
\caption{Plots for left-recursive grammars of increasing size}
\label{fig:lr-grammars}
\end{figure}

\begin{figure}[htp]
\centering
\includegraphics[width=.3\textwidth]{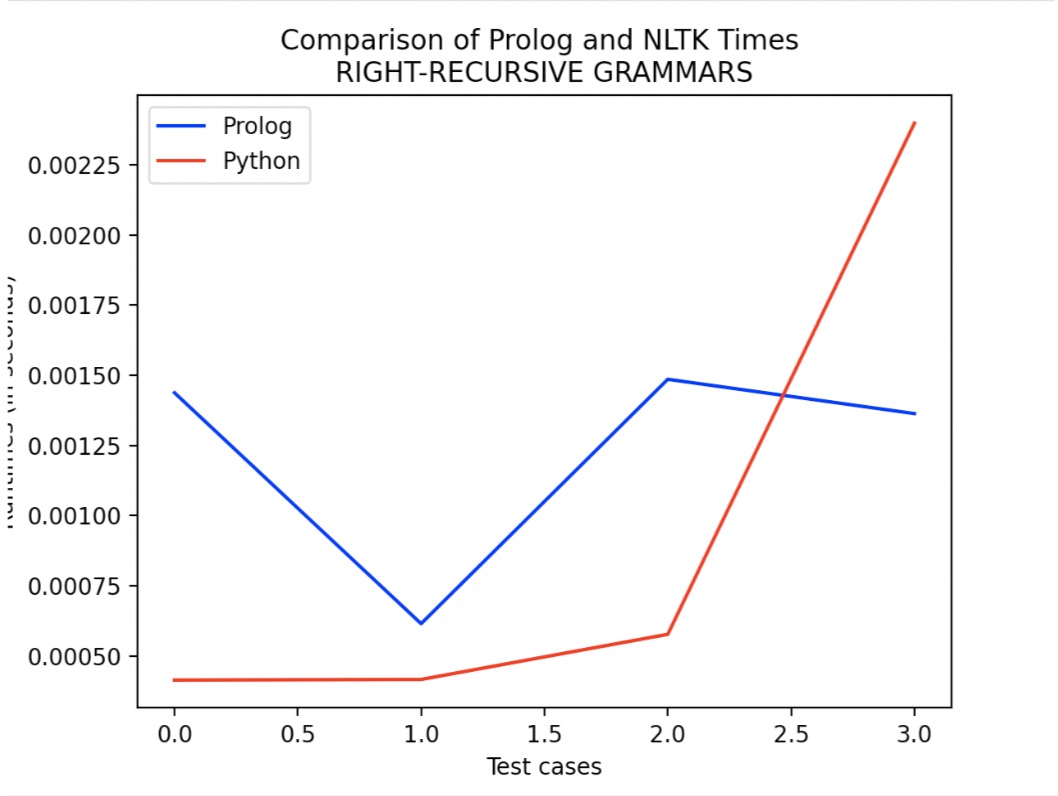}\hfill
\includegraphics[width=.3\textwidth]{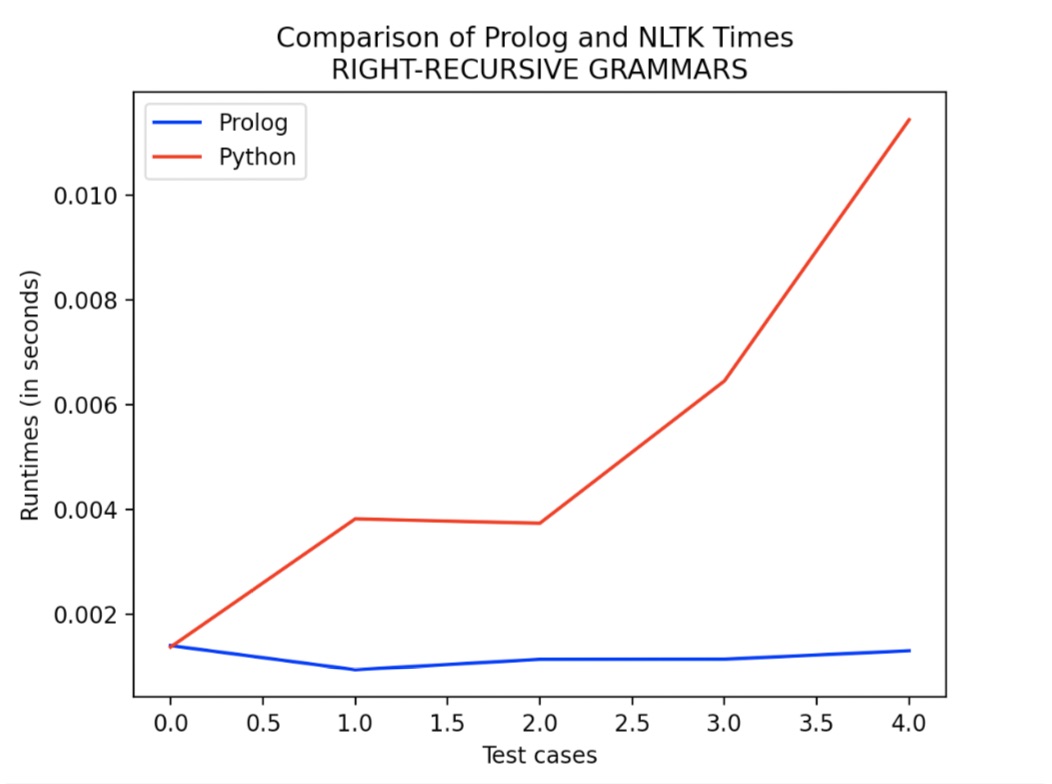}\hfill
\includegraphics[width=.3\textwidth]{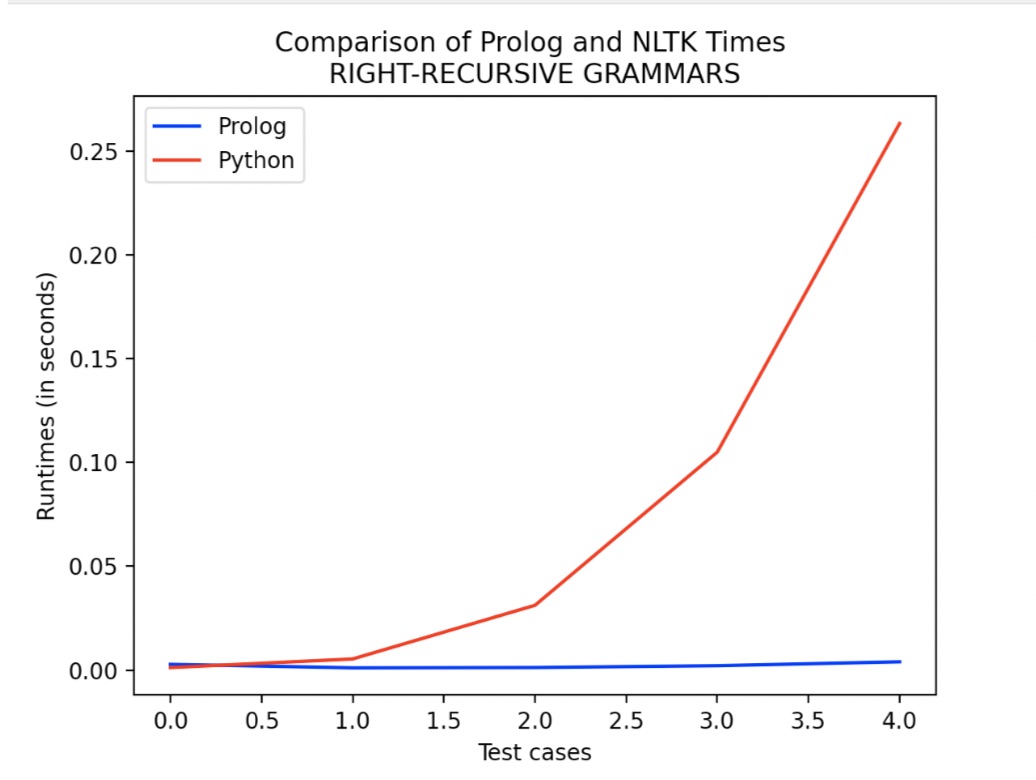}
\caption{Running times 
for right-recursive grammars of increasing size}
\label{fig:rr-grammars}
\end{figure}

\begin{figure}[htp]
\centering
\includegraphics[width=.3\textwidth]{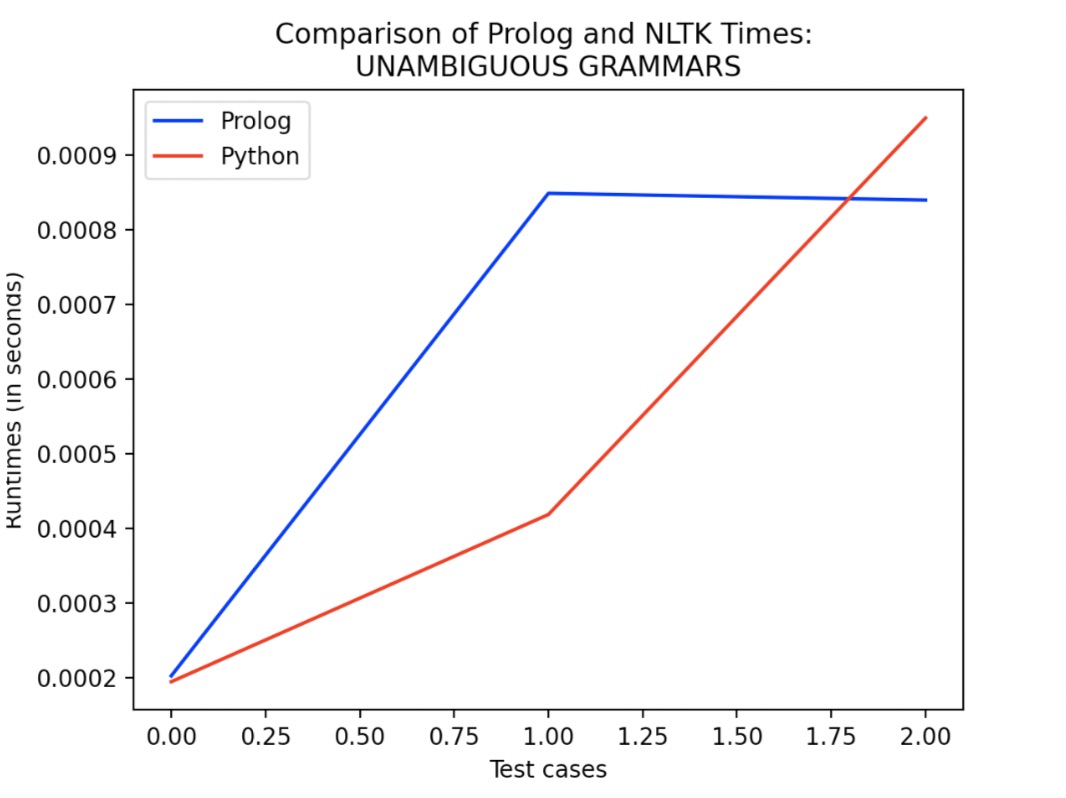}\hfill
\includegraphics[width=.3\textwidth]{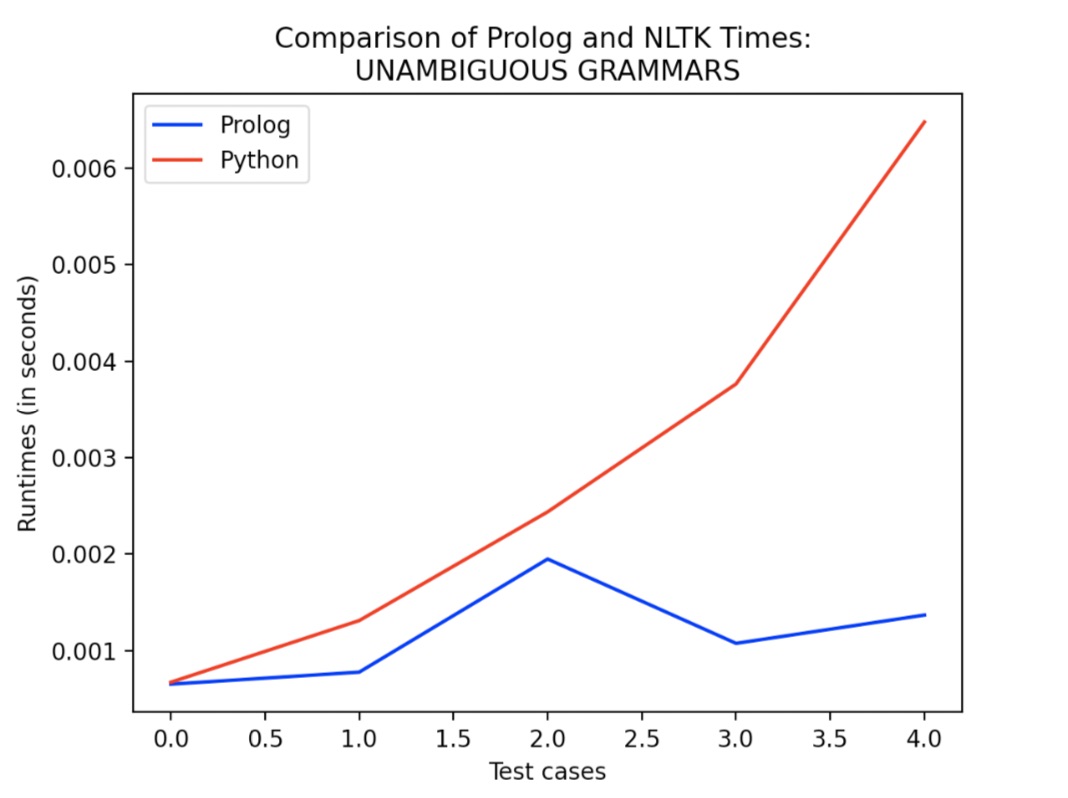}\hfill
\includegraphics[width=.3\textwidth]{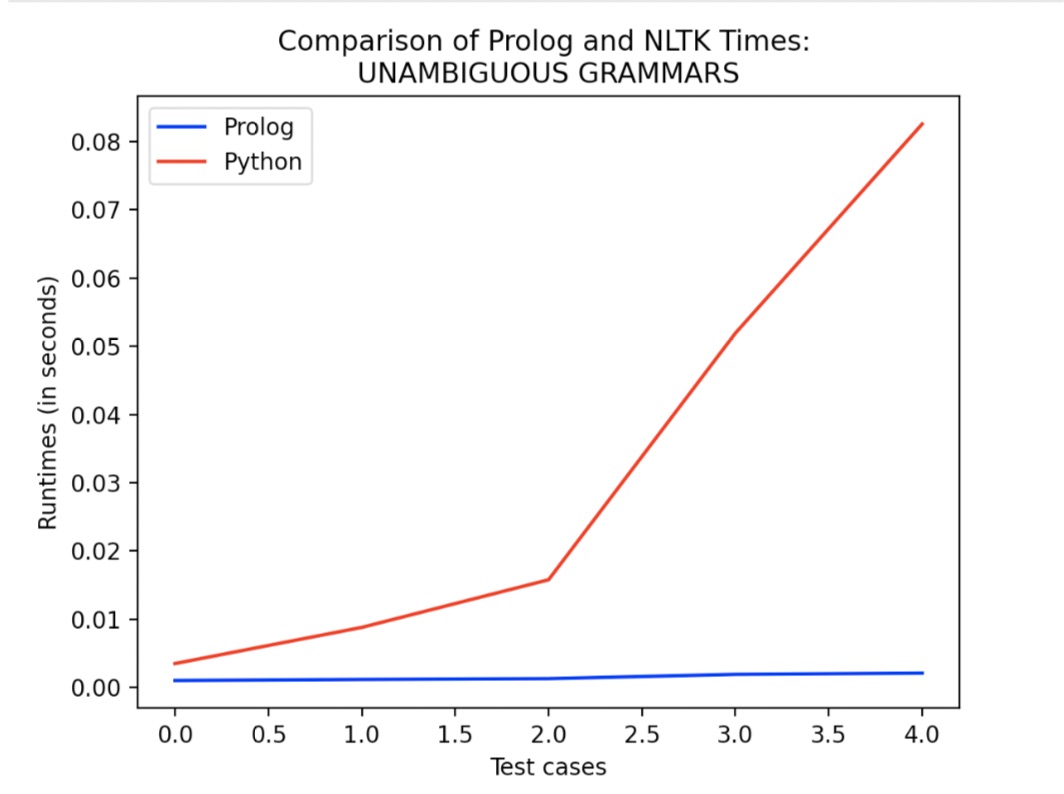}
\caption{Running times for unambiguous grammars of increasing size}
\label{fig:unamb-grammars}
\end{figure}

\begin{figure}[htp]
\centering
\includegraphics[width=.3\textwidth]{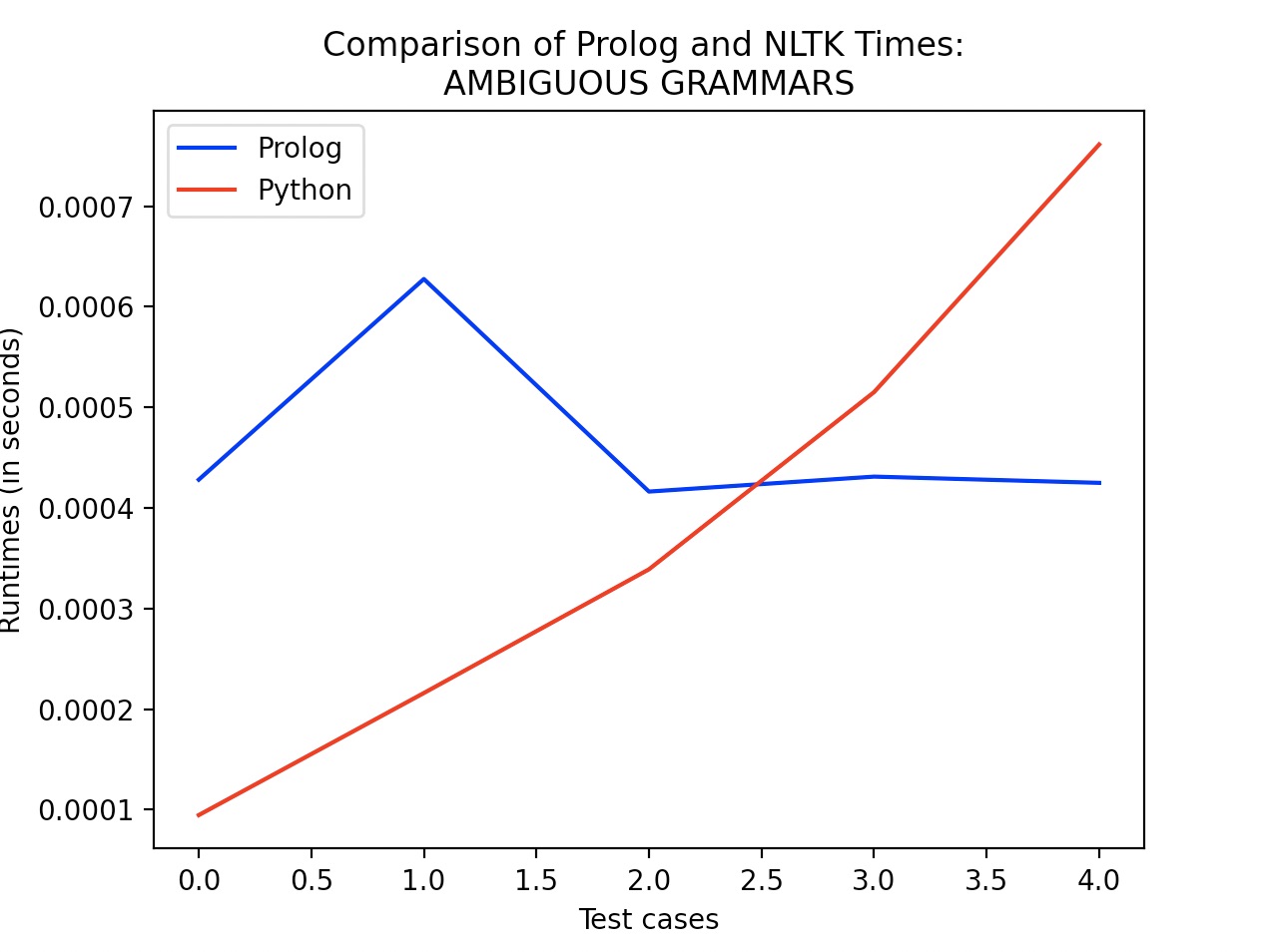}\hfill
\includegraphics[width=.3\textwidth]{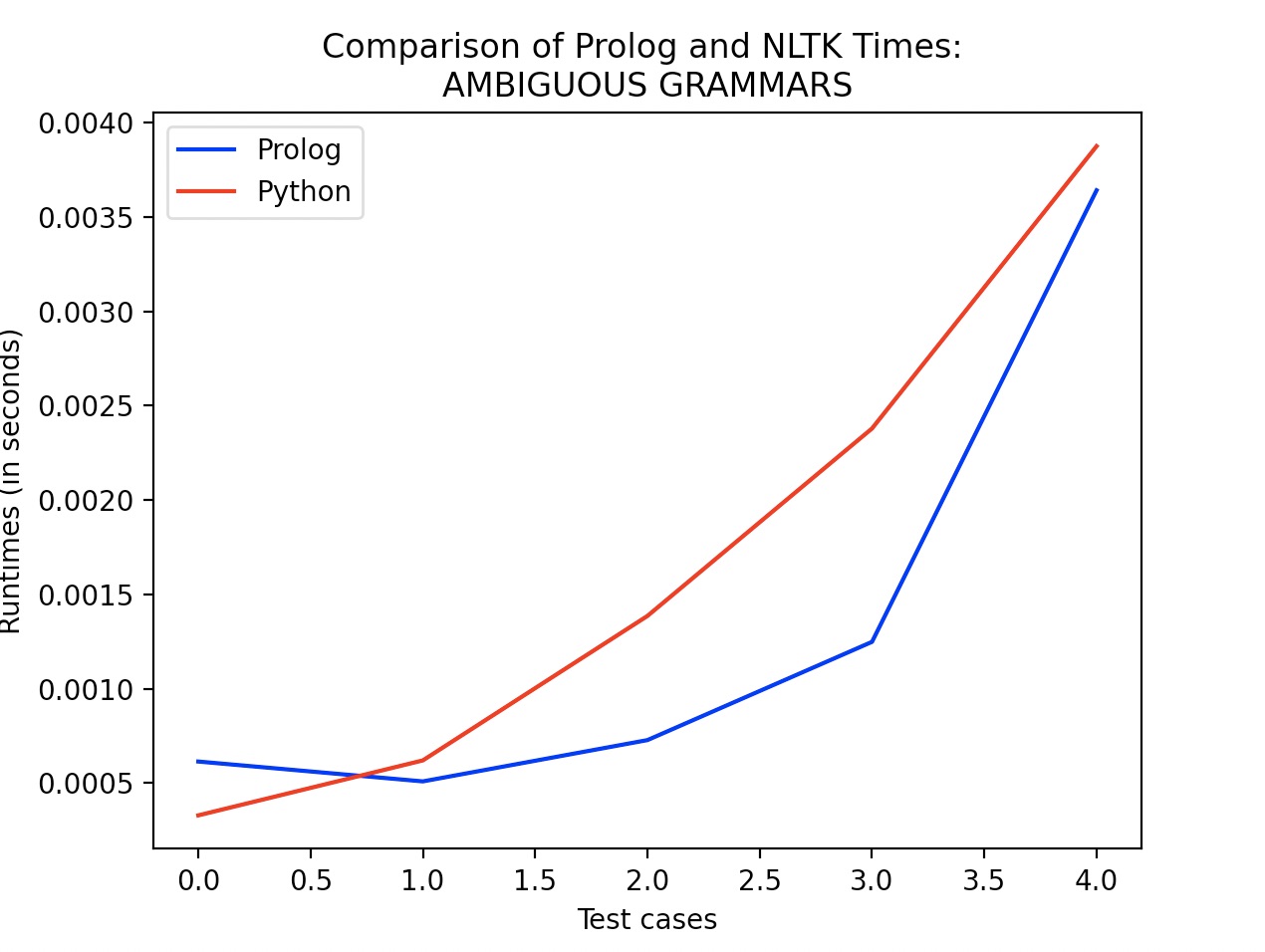}\hfill
\includegraphics[width=.3\textwidth]{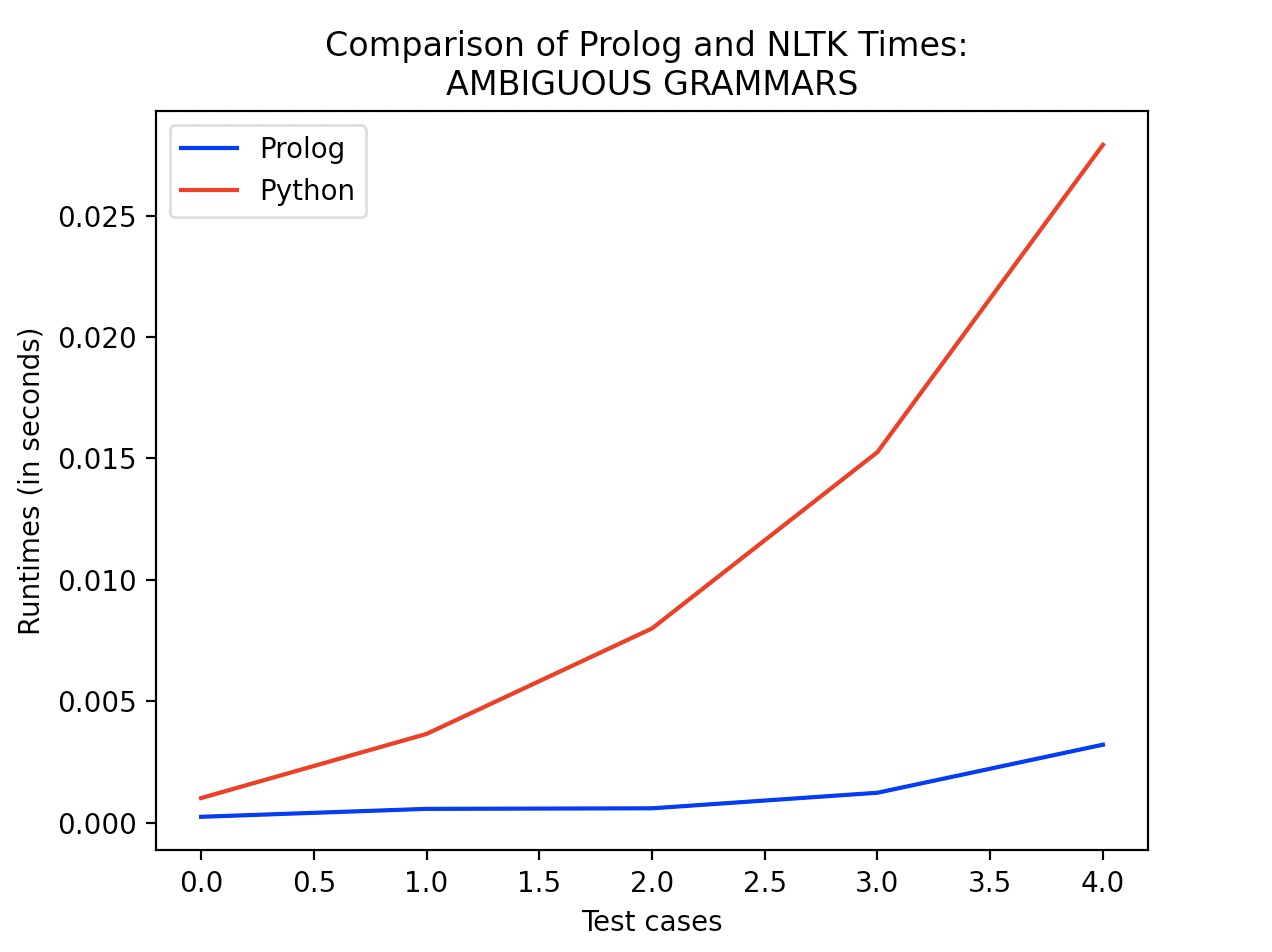}
\caption{Running times for ambiguous grammars of increasing size}
\label{fig:amb-grammars}
\end{figure}

We ran experiments testing 
these DCG parsers
in comparison with the current Viterbi parser API in the Python Natural Language Toolkit (NLTK). The Viterbi algorithm here uses a greedy heuristic, while our parsing algorithm performs an enumeration of all possible parses before choosing the optimal one.
%
Figures~\ref{fig:lr-grammars},~\ref{fig:rr-grammars},~\ref{fig:unamb-grammars}, and~\ref{fig:amb-grammars} show the running times of sentence parses on grammars of increasing size, for each type of grammar.
The x-axis represents the test cases, i.e. each point is a test case, with each test case representing an input sentence ranging from lengths 1 to 50. Higher numbered test cases represent sentences with longer lengths.  
The y-axis is the running time of sentence parse in seconds, averaged over 10 runs.
All measurements were taken on a machine with a 2GHz Quad-Core Intel Core i5 processor, 16GB RAM, running MacOS 14.3.1, with Python 3.11.4 and XSB version 5.0.

Across all types of grammars (left-recursive, right-recursive, unambiguous, ambiguous), the results are uniform: for large grammars with 100+ production rules, i.e. test 3, our Prolog parser runs much more efficiently. In particular, for left-recursive, right-recursive, and unambiguous grammars, our parser is observed to run in linear time in
the length of the input sentence for large grammars.

\mysec{Related work, future work, and conclusion}
\label{sec:related}
The most notable line of work similar to ours is Retrieval Augmented Generation (RAG), an architectural approach that augments LLMs with external knowledge such as databases~\cite{rag-survey}. RAG is particularly useful in knowledge-intensive scenarios or domain-specific applications that require continually updated knowledge; it ensures that the response of an LLM is not based solely on static training data and rather uses up-to-date external data sources to provide responses. RAG has been popularized recently with its application in conversational agents. Our work has the similar motivations as RAG, but we use a ``built-in'' knowledge base to store facts used for context and utilize semantic parsing implemented in XSB Prolog to insert and retrieve information from the KB.

Our work also has similar motivations to that of KALM, a logic system for authoring facts and questions~\cite{kalm}. While KALM uses the answer set programming system DLV as the logical system for reasoning about knowledge, our work uses DCG and tabling in XSB Prolog. But as shown in the work of~\cite{xsb-evaluations} using OpenRuleBench to analyze the performance and scalability of different rule engines including XSB and DLV, XSB exhibits significantly better runtime performance than DLV on various tasks due to tabling.

A limitation to LP-LM is the generalization of English sentences, since we represent the grammar rules as PCFGs manually. Although new grammar rules can always be added at anytime, doing so can be tedious, and there are sentences that intentionally violate grammatical rules or standard sentence structures. In this case, we can simply ``augment'' LP-LM to use LLMs or other NLP techniques for input pre-processing to help extract filler words and distill the core facts from sentences, for example by fine-tuning text summarization models. Regarding the method itself, LP-LM is limited in that the class of queries the system can answer is limited to simple retrieval tasks that do not require any form of reasoning. Getting LP-LM to support reasoning capabilities such as deductive and inductive reasoning, as well as further generalizing the system, are plans for our future work.

In conclusion, while LLMs use deep learning models and are trained on massive datasets, 
making them prone to hallucinations, 
our work, 
LP-LM, 
shows that a KB of facts and a question implemented using Prolog's DCG and tabling for efficient semantics parsing of PCFG can produce reliable answers and produce them efficiently.

{
\renewcommand{\baselinestretch}{-0}
\small

\bibliography{refs}

\begin{thebibliography}{10}
\providecommand{\bibitemdeclare}[2]{}
\providecommand{\surnamestart}{}
\providecommand{\surnameend}{}
\providecommand{\urlprefix}{Available at }
\providecommand{\url}[1]{\texttt{#1}}
\providecommand{\href}[2]{\texttt{#2}}
\providecommand{\urlalt}[2]{\href{#1}{#2}}
\providecommand{\doi}[1]{doi:\urlalt{https://doi.org/#1}{#1}}
\providecommand{\eprint}[1]{arXiv:\urlalt{https://arxiv.org/abs/#1}{#1}}
\providecommand{\bibinfo}[2]{#2}

\bibitemdeclare{article}{uni-grammars-relaxed}
\bibitem{uni-grammars-relaxed}
\bibinfo{author}{Tony \surnamestart Abou-Assaleh\surnameend}, \bibinfo{author}{Nick \surnamestart Cercone\surnameend} \& \bibinfo{author}{Vlado \surnamestart Keselj\surnameend} (\bibinfo{year}{2003}): \emph{\bibinfo{title}{Expressing Probabilistic Context-Free Grammars in the Relaxed Unification Formalism}}.
\newblock {\slshape \bibinfo{journal}{Proceedings of the Conference Pacific Association for Computational Linguistics}}, pp. \bibinfo{pages}{29--36}.

\bibitemdeclare{misc}{chatgpt-qas}
\bibitem{chatgpt-qas}
\bibinfo{author}{Hossein \surnamestart Bahak\surnameend}, \bibinfo{author}{Farzaneh \surnamestart Taheri\surnameend}, \bibinfo{author}{Zahra \surnamestart Zojaji\surnameend} \& \bibinfo{author}{Arefeh \surnamestart Kazemi\surnameend} (\bibinfo{year}{2023}): \emph{\bibinfo{title}{Evaluating ChatGPT as a Question Answering System: A Comprehensive Analysis and Comparison with Existing Models}}, \doi{10.48550/arXiv.2312.07592}.

\bibitemdeclare{inproceedings}{xsb-evaluations}
\bibitem{xsb-evaluations}
\bibinfo{author}{Stefan \surnamestart Brass\surnameend} \& \bibinfo{author}{Heike \surnamestart Stephan\surnameend} (\bibinfo{year}{2017}): \emph{\bibinfo{title}{Experiences with Some Benchmarks for Deductive Databases and Implementations of Bottom-Up Evaluation}}.
\newblock In: {\slshape \bibinfo{booktitle}{Proceedings of the 24th International Workshop on Functional and (Constraint) Logic Programming}}, {\slshape \bibinfo{series}{{EPTCS}}} \bibinfo{volume}{234}, pp. \bibinfo{pages}{57--72}, \doi{10.4204/EPTCS.234.5}.

\bibitemdeclare{book}{pcfg-2}
\bibitem{pcfg-2}
\bibinfo{author}{Glenn \surnamestart Carroll\surnameend} \& \bibinfo{author}{Eugene \surnamestart Charniak\surnameend} (\bibinfo{year}{1992}): \emph{\bibinfo{title}{Two experiments on learning probabilistic dependency grammars from corpora}}.
\newblock \bibinfo{publisher}{Brown University Department of Computer Science}.

\bibitemdeclare{book}{cocke-cyk}
\bibitem{cocke-cyk}
\bibinfo{author}{John \surnamestart Cocke\surnameend} (\bibinfo{year}{1969}): \emph{\bibinfo{title}{Programming languages and their compilers: Preliminary notes}}.
\newblock \bibinfo{publisher}{New York University}.

\bibitemdeclare{article}{earley}
\bibitem{earley}
\bibinfo{author}{Jay \surnamestart Earley\surnameend} (\bibinfo{year}{1970}): \emph{\bibinfo{title}{An efficient context-free parsing algorithm}}.
\newblock {\slshape \bibinfo{journal}{Commun. ACM}} \bibinfo{volume}{13}(\bibinfo{number}{2}), p. \bibinfo{pages}{94–102}, \doi{10.1145/362007.362035}.

\bibitemdeclare{misc}{rag-survey}
\bibitem{rag-survey}
\bibinfo{author}{Yunfan \surnamestart Gao\surnameend}, \bibinfo{author}{Yun \surnamestart Xiong\surnameend}, \bibinfo{author}{Xinyu \surnamestart Gao\surnameend}, \bibinfo{author}{Kangxiang \surnamestart Jia\surnameend}, \bibinfo{author}{Jinliu \surnamestart Pan\surnameend}, \bibinfo{author}{Yuxi \surnamestart Bi\surnameend}, \bibinfo{author}{Yi~\surnamestart Dai\surnameend}, \bibinfo{author}{Jiawei \surnamestart Sun\surnameend}, \bibinfo{author}{Meng \surnamestart Wang\surnameend} \& \bibinfo{author}{Haofen \surnamestart Wang\surnameend} (\bibinfo{year}{2024}): \emph{\bibinfo{title}{Retrieval-Augmented Generation for Large Language Models: A Survey}}, \doi{10.48550/arXiv.2312.10997}.

\bibitemdeclare{article}{stochastic-dcg}
\bibitem{stochastic-dcg}
\bibinfo{author}{Christian~Theil \surnamestart Have\surnameend} (\bibinfo{year}{2009}): \emph{\bibinfo{title}{Stochastic definite clause grammars}}.
\newblock {\slshape \bibinfo{journal}{Proceedings of the International Conference RANLP}}, pp. \bibinfo{pages}{139--143}.
\newblock \urlprefix\url{https://aclanthology.org/R09-1027/}.

\bibitemdeclare{article}{hallucination-survey}
\bibitem{hallucination-survey}
\bibinfo{author}{Ziwei \surnamestart Ji\surnameend}, \bibinfo{author}{Nayeon \surnamestart Lee\surnameend}, \bibinfo{author}{Rita \surnamestart Frieske\surnameend}, \bibinfo{author}{Tiezheng \surnamestart Yu\surnameend}, \bibinfo{author}{Dan \surnamestart Su\surnameend}, \bibinfo{author}{Yan \surnamestart Xu\surnameend}, \bibinfo{author}{Etsuko \surnamestart Ishii\surnameend}, \bibinfo{author}{Ye~Jin \surnamestart Bang\surnameend}, \bibinfo{author}{Andrea \surnamestart Madotto\surnameend} \& \bibinfo{author}{Pascale \surnamestart Fung\surnameend} (\bibinfo{year}{2023}): \emph{\bibinfo{title}{Survey of Hallucination in Natural Language Generation}}.
\newblock {\slshape \bibinfo{journal}{ACM Computing Surveys}} \bibinfo{volume}{55}(\bibinfo{number}{12}), p. \bibinfo{pages}{1–38}, \doi{10.1145/3571730}.

\bibitemdeclare{techreport}{kasami-cyk}
\bibitem{kasami-cyk}
\bibinfo{author}{Tadao \surnamestart Kasami\surnameend} (\bibinfo{year}{1965}): \emph{\bibinfo{title}{An efficient recognition and syntax-analysis algorithm for context-free languages}}.
\newblock \bibinfo{type}{Technical Report}, \bibinfo{institution}{Air Force Cambridge Research Lab, Bedford, MA}.

\bibitemdeclare{article}{pcfg-1}
\bibitem{pcfg-1}
\bibinfo{author}{Karim \surnamestart Lari\surnameend} \& \bibinfo{author}{Steve~J. \surnamestart Young\surnameend} (\bibinfo{year}{1990}): \emph{\bibinfo{title}{The estimation of stochastic context-free grammars using the inside-outside algorithm}}.
\newblock {\slshape \bibinfo{journal}{Computer speech \& language}} \bibinfo{volume}{4}(\bibinfo{number}{1}), pp. \bibinfo{pages}{35--56}, \doi{10.1016/0885-2308(90)90022-X}.

\bibitemdeclare{inproceedings}{SagSW94xsb}
\bibitem{SagSW94xsb}
\bibinfo{author}{Konstantinos \surnamestart Sagonas\surnameend}, \bibinfo{author}{Terrance \surnamestart Swift\surnameend} \& \bibinfo{author}{David~S. \surnamestart Warren\surnameend} (\bibinfo{year}{1994}): \emph{\bibinfo{title}{{XSB} as an Efficient Deductive Database Engine}}.
\newblock In: {\slshape \bibinfo{booktitle}{Proceedings of the 1994 ACM SIGMOD International Conference on Management of Data}}, \bibinfo{publisher}{ACM}, pp. \bibinfo{pages}{442--453}, \doi{10.1145/191839.191927}.

\bibitemdeclare{inproceedings}{shift-reduce}
\bibitem{shift-reduce}
\bibinfo{author}{Seppo \surnamestart Sippu\surnameend} \& \bibinfo{author}{Eljas \surnamestart Soisalon-Soininen\surnameend} (\bibinfo{year}{1988}): \emph{\bibinfo{title}{Parsing Theory - Volume I: Languages and Parsing}}.
\newblock In: {\slshape \bibinfo{booktitle}{EATCS Monographs on Theoretical Computer Science}}, \doi{10.1007/978-3-642-61345-6}.

\bibitemdeclare{article}{uni-grammars-prob}
\bibitem{uni-grammars-prob}
\bibinfo{author}{Tony~C. \surnamestart Smith\surnameend} \& \bibinfo{author}{John~G. \surnamestart Cleary\surnameend} (\bibinfo{year}{1997}): \emph{\bibinfo{title}{Probabilistic unification grammars}}.
\newblock {\slshape \bibinfo{journal}{Australasian Natural Language Processing Summer Workshop}}.

\bibitemdeclare{manual}{xsb22}
\bibitem{xsb22}
\bibinfo{author}{Theresa \surnamestart Swift\surnameend}, \bibinfo{author}{David~S. \surnamestart Warren\surnameend}, \bibinfo{author}{Konstantinos \surnamestart Sagonas\surnameend}, \bibinfo{author}{Juliana \surnamestart Freire\surnameend}, \bibinfo{author}{Prasad \surnamestart Rao\surnameend}, \bibinfo{author}{Baoqiu \surnamestart Cui\surnameend}, \bibinfo{author}{Ernie \surnamestart Johnson\surnameend}, \bibinfo{author}{Luis \surnamestart de~Castro\surnameend}, \bibinfo{author}{Rui~F. \surnamestart Marques\surnameend}, \bibinfo{author}{Diptikalyan \surnamestart Saha\surnameend}, \bibinfo{author}{Steve \surnamestart Dawson\surnameend} \& \bibinfo{author}{Michael \surnamestart Kifer\surnameend} (\bibinfo{year}{2022}): \emph{\bibinfo{title}{The XSB System Version 5.0,x}}.
\newblock \bibinfo{note}{\url{http://xsb.sourceforge.net}. Latest release May 12, 2022.}

\bibitemdeclare{inproceedings}{kalm}
\bibitem{kalm}
\bibinfo{author}{Yuheng \surnamestart Wang\surnameend}, \bibinfo{author}{Paul \surnamestart Fodor\surnameend} \& \bibinfo{author}{Michael \surnamestart Kifer\surnameend} (\bibinfo{year}{2023}): \emph{\bibinfo{title}{Knowledge Authoring for Rules and Actions}}.
\newblock In: {\slshape \bibinfo{booktitle}{International Conference on Logic Programming}}, \bibinfo{series}{{TPLP}}, \doi{10.1017/S1471068423000169}.

\bibitemdeclare{article}{younger-cyk}
\bibitem{younger-cyk}
\bibinfo{author}{Daniel~H. \surnamestart Younger\surnameend} (\bibinfo{year}{1967}): \emph{\bibinfo{title}{Recognition and parsing of context-free languages in time n3}}.
\newblock {\slshape \bibinfo{journal}{Information and Control}} \bibinfo{volume}{10}(\bibinfo{number}{2}), pp. \bibinfo{pages}{189--208}, \doi{10.1016/S0019-9958(67)80007-X}.

\end{thebibliography}
\bibliographystyle{eptcs}
}

\end{document}